\pdfoutput=1

\documentclass[11pt]{article}

\usepackage{EMNLP2023}

\usepackage{times}
\usepackage{latexsym}

\usepackage[T1]{fontenc}

\usepackage[utf8]{inputenc}

\usepackage{microtype}

\usepackage{inconsolata}

\usepackage{graphicx}
\usepackage{multirow}
\usepackage{multicol}
\usepackage{array}
\usepackage{xurl}

\usepackage{enumitem}

\title{Data-to-text Generation for Severely Under-Resourced Languages\\with GPT-3.5: A Bit of Help Needed from Google Translate}

\author{Michela Lorandi \and Anya Belz \\
  ADAPT Research Centre\\
  Dublin City University,
  Ireland\\
  \texttt{\{michela.lorandi,anya.belz\}@adaptcentre.ie}}

\begin{document}
\maketitle
\begin{abstract}
LLMs like GPT are great at tasks involving English which dominates in their training data.
In this paper, we look at how they cope with tasks involving languages that are severely under-represented in their training data, in the context of data-to-text generation for Irish, Maltese, Welsh and Breton.
During the prompt-engineering phase we tested a range of prompt types and formats on GPT-3.5 and~4 with a small sample of example input/output pairs. We then fully evaluated the two most promising prompts in two scenarios: (i) direct generation into the under-resourced language, and (ii) generation into English followed by translation into the under-resourced language.
We find that few-shot prompting works better for direct generation into under-resourced languages, but that the difference disappears when  pivoting via English. The few-shot + translation system variants were submitted to the WebNLG 2023 shared task where they outperformed competitor systems by substantial margins in all languages on all metrics. 
We conclude that good performance on under-resourced languages can be achieved out-of-the box with state-of-the-art LLMs. However, our best results (for Welsh)
remain well below the lowest ranked English system at WebNLG'20.
\end{abstract}

\section{Introduction}\label{sec:intro}
\vspace{-.1cm}

Pretrained large language models (LLMs) are everywhere, achieving state-of-the-art-results across many application areas. LLMs  display strong out-of-the-box abilities at a wide range of different tasks far beyond language-based tasks. However, these abilities have  been explored mostly for  English  which dominates text online, hence the data on which LLMs are trained. In the work reported here, we explore to what  extent  state-of-the-art LLMs can help with tasks involving severely under-resourced languages which by definition would have been very scarce in their training data.

More specifically, we address the challenging task of data-to-text generation for four severely under-resourced languages: Irish, Maltese, Welsh and Breton. Developing any new system is difficult in a severely under-resourced scenario, more so where the task has language-specific input/output combinations as in data-to-text and text-to-text.

The paper is structured as follows. Section~\ref{sec:data-task} describes data and task, Section~\ref{sec:system} presents the general approach,  prompt engineering, and the specific systems we  fully evaluated. Experimental set-up and results are outlined in Section~\ref{sec:exp-setup}, and  Section~\ref{sec:disc} provides discussion and conclusions.

All code and results are available on Github: \url{https://github.com/DCU-NLG/DCU-NLG-PBN}.

\section{Data and Task}
\label{sec:data-task}
\vspace{-.1cm}

The WebNLG'23 \cite{aquilina-etal-2023} data consists of 1,778 dev items
for each language,  1,399 test items for Breton, and 1,665 for Welsh, Irish and Maltese, that were manually translated by professional translators from the English originals. Additionally 13,211 training items are provided where  texts were automatically translated from English.  

\begin{figure*}[h!tb]
\vspace{-1cm}
\centering
    \includegraphics[width=.8\textwidth]{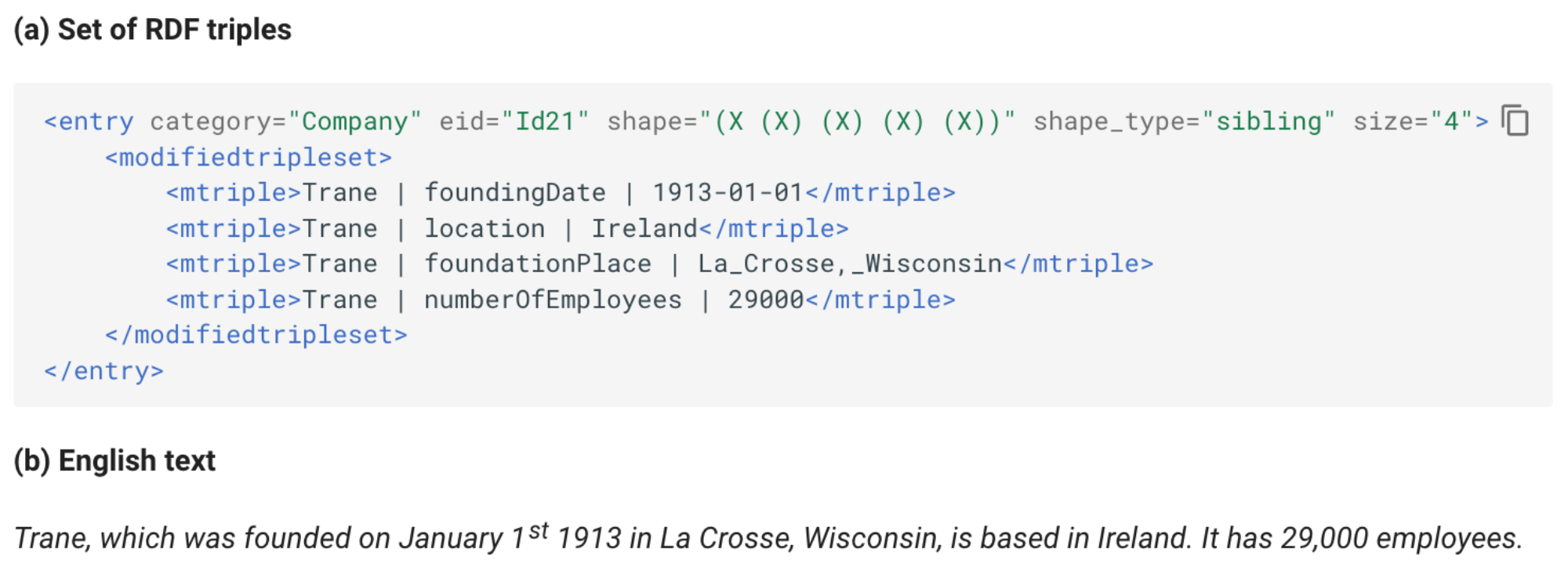}
    \caption{WebNLG example of input triples and output text in English.}
    \label{fig:example}
\end{figure*}

As in all WebNLG shared tasks, WebNLG 2023 systems must map from the data (RDF triples) to a suitable output text, as in the example from the WebNLG'23 website\footnote{\url{https://synalp.gitlabpages.inria.fr/webnlg-challenge/docs}}  in Figure~\ref{fig:example}. This is a data-to-text generation task which is well established in the natural language generation (NLG) field.

\section{General Approach and Specific Systems}
\label{sec:system}
\vspace{-.1cm}

The general approach we take is to use a state-of-the-art LLM in its out-of-the-box state, to either (i) generate texts directly from the data in the target language, or (ii) generate  texts in English and then translate them into the target language. More specifically, the LLM we used was the InstructGPT model identified as text-davinci-003, and  
the MT system was Google Translate which covers all WebNLG'23 languages except Breton.

In all cases, we fed a prompt representing the generation task to text-davinci-003, and collected the system completion as the output which we then cleaned up by simple post-processing procedures.

\begin{figure*}[t!]
\includegraphics[width=\linewidth]{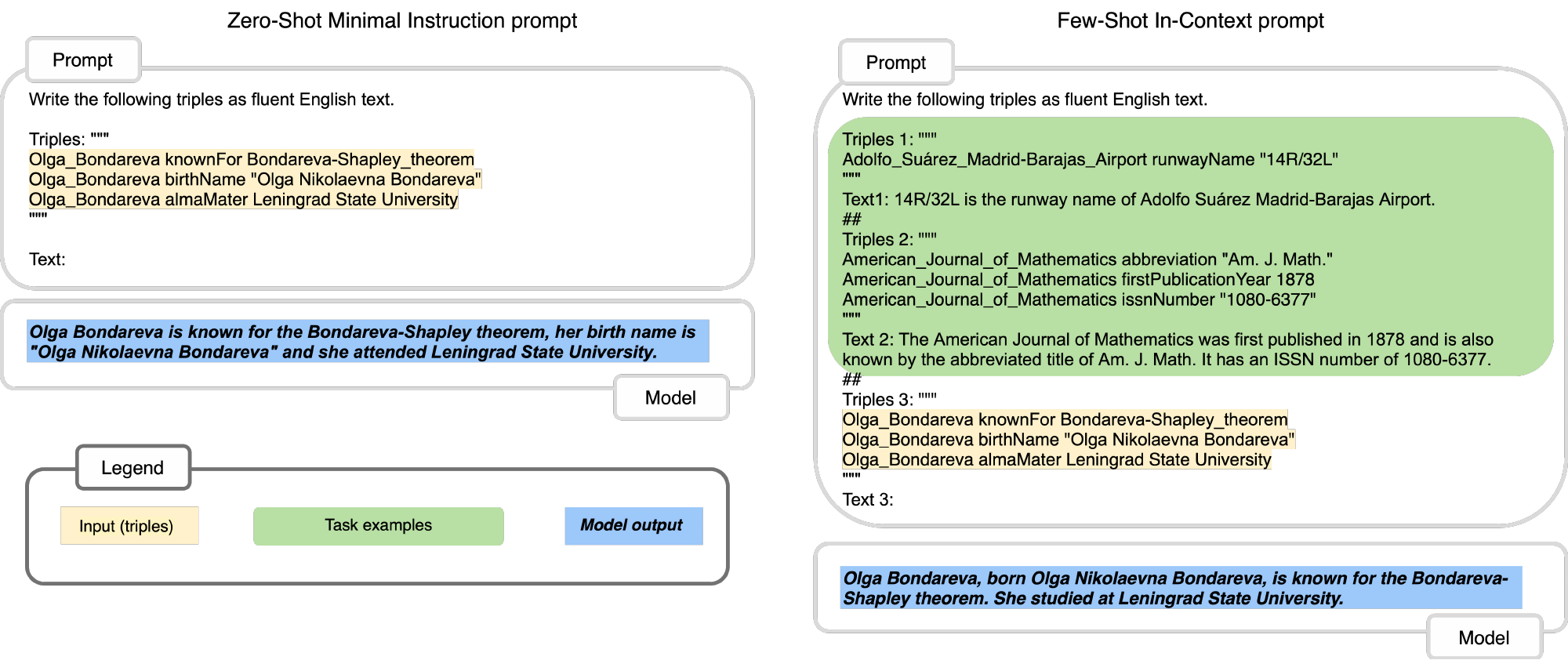}
\centering
\caption{The two  prompt types used in the final systems. Given {Input} (triples) highlighted in yellow,  model output in blue. The few-shot in-context prompt also incorporates examples (highlighted green).\footnotemark}
\label{fig:prompt_types}
\end{figure*}

\subsection{Prompt testing and selection}

We conducted three prompt testing phases, in each phase using a different random sample of 20 data/text pairs, stratified for triple size, with outputs in all target languages, from the dev set.
These small subsets of the dev set represent all the data we used during system development. 

\textbf{Phase 1:} In the first prompt testing phase, we tested seven prompt variants on text-davinci-003 on all four target languages and a subset of four prompts on GPT-4 on Irish and Breton. The prompts were of three types: \textbf{zero-shot minimal instruction}, \textbf{few-shot in-context learning}, and \textbf{chain-of-thought} \cite{wei2022chain}. 

On GPT-3.5, we tested each prompt in two scenarios: (i) generating the texts directly into the target languages, (ii) generating all texts into English and then translating them into each target language. 
In all prompts, the input triples were surrounded by triple quotes as per best-practice recommendations; we tested removing these quotes for the zero-shot  prompt. On GPT-4, we tested zero-shot  and few-shot  with and without translation, and chain-of-thought (CoT) without translation.

The clearest results immediately were that (i) generation into English plus translation worked better than direct generation into the target languages, and (ii) the CoT prompts performed far worse than the others, a difference that was exacerbated when using GPT-4. Zero-shot looked better for Irish, but few-shot had the edge for the other three languages.  

BLEU, METEOR and ChrF++ for zero-shot looked overall slightly better on GPT-4, but TER was substantially worse and moreover the emerging patterns were less clear. Leaving out the triple quotes  did not appear to work better (although this was tested on only one language).

\textbf{Phase 2:} As we were testing on a very small sample, the risk was poor representativeness of the data as a whole. Therefore, in the second prompt testing phase, we used a fresh sample of 20 pairs to further test the most promising  zero-shot and few-shot prompts in both +/- translation scenarios. 
We dropped the chain-of-thought prompts and GPT-4, as the Phase~1 results had been less promising.

Results confirmed better performance for few-shot, this time in all scenarios with all evaluation metrics (although for a small number of results, scores were virtually the same). Again having quotes worked better than leaving them out.

At this point we tested a postprocessing method on most of the previous experiments. This revealed clear improvements in almost all cases. The method replaces HTML with ASCII characters (e.g. \&quot is replaced with "). Furthermore, we remove underscores, and quotes that wrap the entire text.

\textbf{Phase 3:} In the third prompt testing phase, we tested just the zero-shot and few-shot prompts into English with translation, on Irish and Welsh. At this stage the aim was to further investigate the difference between the two so far most promising types of prompts, on another fresh data sample, on these two languages because results were most similar for different prompt variants for them. However, Phase~3 results for zero-shot and few-shot + MT for Irish and Welsh were again overall tied.

\subsection{Prompt types used in final systems}
\label{sec:prompt_types}

Prompt testing had revealed zero-shot and few-shot prompts with triple quotes  generating into English plus MT as clearly the most promising. Figure~\ref{fig:prompt_types} shows examples of these two prompt variants in terms of the prompt template, few-shot examples (where applicable), and system completion. For more example prompts, see appendix, Section~\ref{sec:appendix_prompts}. \textbf{Zero-shot minimal instruction} prompts are simple requests to perform a task without details or examples, in our case consisting of a brief sentence describing the task and the input to the task. \textbf{Few-shot in-context learning} prompts are composed of a brief description of the task and a list of examples providing both inputs and target outputs, ending in  just an input, for which the system response is taken to be the output. 

The final prompt formats follow best practice for prompt engineering provided by OpenAI;\footnote{\url{https://help.openai.com/en/articles/6654000-best-practices-for-prompt-engineering-with-openai-api}} input triples are recast as \textit{subject predicate object}.

\subsection{System variants evaluated}

In the full evaluations on the WebNLG'23 test set reported below, we tested our selected zero-shot and few-shot prompts in both +/- MT scenarios, giving us four system variants as follows:

\begin{enumerate}[itemsep=-1pt]
    \item Zero-shot generation into each language;
    \item Few-shot generation into each language;
    \item Zero-shot generation into English + translation into each language;
    \item Few-shot generation into English + translation into each language.
\end{enumerate}

\section{Experimental Set-up and Results}
\label{sec:exp-setup}
\vspace{-.1cm}

We executed our experiments using the paid-for OpenAI API to access text-davinci-003,\footnote{\url{https://platform.openai.com/docs/models/gpt-3-5}} and  the free Google Translate API,\footnote{\url{https://cloud.google.com/translate}} in early June 2023. 

In all experiments, we set text-davinci-003 parameters to \textit{temperature}=0, \textit{top p}=1 (default) to consider all possible tokens, \textit{frequency penalty}=0 and \textit{presence penalty}=0 (default value) to slightly reduce repetitive tokens, \textit{best of}=1 (default) to get only 1 completion for each prompt, and \textit{maximum length} to 500. All generated texts are post-processed as described above. 

\begin{table*}[h!]
    \centering
    \small
    \setlength\tabcolsep{10pt} 
    \renewcommand{\arraystretch}{1.15}
    \begin{tabular}{|l|l|ccc|}
        \hline
        \textbf{Language} & \textbf{Method} & \textbf{BLEU} $\uparrow$ & \textbf{ChrF++} $\uparrow$ & \textbf{TER} $\downarrow$ \\
        \hline
        \multirow{5}{*}{Irish} & Zero-Shot Irish & 12.9931 & 0.4124 & 0.9298 \\
         & Few-Shot Irish & 15.3477 & 0.4303 & 0.8451 \\
         & Zero-Shot English + Google Translate & \textbf{20.5176} & \textbf{0.5146} & 0.7122 \\
         & Few-Shot English + Google Translate & 20.4001 & 0.51 & \textbf{0.6894} \\ \cline{2-5}
         & WebNLG 2023 Baseline & 11.63 & 0.36 & 0.74 \\
        \hline
        \hline
        \multirow{5}{*}{Maltese} & Zero-Shot Maltese & 13.0311 & 0.445 & 0.8496 \\
         & Few-Shot Maltese & 15.4315 & 0.4536 & 0.7605 \\
         & Zero-Shot English + Google Translate & 20.3528 & \textbf{0.5263} & 0.67 \\
         & Few-Shot English + Google Translate & \textbf{21.2656} & 0.5249 & \textbf{0.6465} \\ \cline{2-5}
         & WebNLG 2023 Baseline & 15.60 & 0.42 & 0.67 \\
        \hline
        \hline
        \multirow{5}{*}{Welsh} & Zero-Shot Welsh & 15.8695 & 0.4619 & 0.822 \\
         & Few-Shot Welsh & 18.9512 & 0.4742 & 0.7192 \\
         & Zero-Shot English + Google Translate & 24.7126 & \textbf{0.5496} & 0.6659 \\
         & Few-Shot English + Google Translate & \textbf{25.115} & 0.5484 & \textbf{0.6435} \\ \cline{2-5}
         & WebNLG 2023 Baseline & 10.70 & 0.36 & 0.77 \\
        \hline
        \hline
        \multirow{3}{*}{Breton} & Zero-Shot Breton & 15.2498 & 0.4217 & 0.82 \\
         & Few-Shot Breton & \textbf{17.207} & \textbf{0.4394} & \textbf{0.739} \\ \cline{2-5}
         & WebNLG 2023 Baseline & 9.92 & 0.33 & 0.76 \\ 
        \hline
    \end{tabular}
    \caption{Results for our system variants (Section~\ref{sec:system}), and the WebNLG'23 baseline results  provided by the organisers. Highest score in each column for each language in boldface.}
\label{tab:prompts_results}
\end{table*}

Table~\ref{tab:prompts_results} presents BLEU \cite{papineni2002bleu}, ChrF++ \cite{popovic-2017-chrf} and TER \cite{snover-etal-2006-study} for our four systems,
and the baseline reported by the WebNLG organisers \cite{aquilina-etal-2023}: the winning system for English from  WebNLG'20 combined with the Zero MT system. 

Comparing  zero-shot results with  corresponding few-shot results, it is clear
that the addition of examples to the prompt 
helps the model  perform the task better in all languages: BLEU and TER are always higher for the few-shot variant than the zero-shot variant, while ChrF++ is always higher for few-shot where translation is \textit{not} used, and virtually identical where translation \textit{is} used.

When generating the texts in English plus translation, we don't see the same strong benefit achieved by the addition of few-shot examples noted above. Even where there is a benefit, it is small. In fact, we observe similar performance between zero-shot + MT and few-shot + MT throughout, while both clearly outperform  versions without translation; this is the case in all three languages where we have translation (Irish, Maltese, and Welsh).

It might be speculated that  providing few-shot examples partially compensates for the comparatively smaller amount of knowledge that the LLM has about the under-resourced languages. This would explain why the benefit disappears when pivoting via English, where examples can add little to the knowledge the LLM already has.

\vspace{-.05cm}
\section{Discussion and Conclusion}
\label{sec:disc}
\vspace{-.1cm}

We submitted the three few-shot + MT systems to the WebNLG Challenge 2023,\footnote{\url{https://synalp.gitlabpages.inria.fr/webnlg-challenge/challenge\_2023/docs}} in the Welsh, Irish and Maltese tracks. 
Based on the metric evaluation results (test set) shared by WebNLG organisers, these systems 
outperform all other submitted systems by sizeable margins; moreover, few-shot Breton without translation ($^*$) also outperforms the one submitted system substantially:

\begin{small}
    \setlength\tabcolsep{2.6pt} 
    \renewcommand{\arraystretch}{1.15}
\begin{tabular}{|l||c|c|c||c|c|c|}
\hline
            & \multicolumn{3}{c||}{Few-Shot + GTransl} & \multicolumn{3}{c|}{Nearest system} \\
\hline
            & BLEU  & chrF++ & TER  & BLEU  & chrF++ & TER  \\
\hline
    Irish & 20.4 & 0.51 & 0.69 & 16.66 & 0.44   & 0.75 \\
    Maltese & 21.27 & 0.52 & 0.65 & 16.49 & 0.47   & 0.7 \\
    Welsh & 25.12 & 0.55 & 0.64 & 20.97 & 0.49   & 0.67 \\
        Breton$^*$ & 17.21 & 0.44 & 0.74 & 13.11 & 0.33 & 0.76 \\

\hline
\end{tabular}    
\end{small}

\noindent At the time of writing, we do not know what kinds of systems produced the runner-up results, except that the Irish runner-up is the  handcrafted rule-based FORGe system extended for Irish \cite{milleWebNLG}. There are interesting differences between FORGe and our system. FORGe took a considerable amount of person time, but uses no paid-for resources. To create and test our system took next to no time, and US\$91.82 
in API costs.
For FORGe, the system and how it was created is known. 
For our system, we don't have access to the model itself, and only incomplete knowledge about its exact technical details at the particular time we access it; on another day, results may differ. This makes experiments non-repeatable, and generalising to future use of the same approach difficult. Moreover, it's not a system that can be deployed in this form, because of the changing nature of what's at the other end of the LLM and MT APIs. 

This raises important general questions about (i) how scientific research is to be conducted with LLMs served as-a-service that deliver best performance, but whose technical details aren't known and moreover change over time; and (ii) how such research results can be utilised in developing systems for real-world deployment. 

In this paper we have explored different prompt types and formats for data-to-text generation for under-resourced languages, with out-of-the-box LLMs with and without machine translation. We found that few-shot + MT achieves new state-of-the-art metric performance at the WebNLG'23 RDF-to-text generation task for the severely under-resourced languages Irish, Maltese, Welsh and Breton. However, given the caveats above, results presented here must  be treated as a rough indication of what is achievable with an LLM+MT approach. 

\section*{Acknowledgements}
The cost of accessing the GPT API was covered by financial support from the DCU-NLG Research Group at DCU. Michela Lorandi's work was conducted with the financial support of the Science Foundation Ireland Centre for Research Training in Digitally-Enhanced Reality (d-real) under Grant No. 18/CRT/6224. Both authors benefit from being members of the ADAPT SFI Research Centre at Dublin City University, funded by the Science Foundation Ireland under Grant Agreement No. 13/RC/2106\_P2. For the purpose of Open Access, the author has applied a CC BY public copyright licence to any Author Accepted Manuscript version arising from this submission.

\bibliography{main}

\begin{thebibliography}{6}
\expandafter\ifx\csname natexlab\endcsname\relax\def\natexlab#1{#1}\fi

\bibitem[{Aquilina et~al.(2023)Aquilina, Belz, Borg, Cripwell, Gardent, Gatt,
  Judge, Lorandi, Nikiforoskaya, Soto-Martinez, and
  Thomson}]{aquilina-etal-2023}
Enrico Aquilina, Anya Belz, Claudia Borg, Liam Cripwell, Claire Gardent, Albert
  Gatt, John Judge, Michela Lorandi, Anna Nikiforoskaya, William Soto-Martinez,
  and Craig Thomson. 2023.
\newblock The 2023 webnlg shared task on low resource languages overview and
  evaluation results (webnlg 2023).
\newblock In \emph{Proceedings of the Workshop on Multimodal, Multilingual
  Natural Language Generation and Multilingual WebNLG Challenge}, Prague, Czech
  Republic.

\bibitem[{Mille et~al.(2023)Mille, U'i~Dhonnchadha, Dasiopoulou, Cassidy,
  Davis, and Belz}]{milleWebNLG}
Simon Mille, Elaine U'i~Dhonnchadha, Stamatia Dasiopoulou, Lauren Cassidy,
  Brian Davis, and Anya Belz. 2023.
\newblock {DCU}/{TCD}-{FORG}e at {W}eb{NLG}’23: {I}rish rules!
\newblock In \emph{Proceedings of the Workshop on Multimodal, Multilingual
  Natural Language Generation and Multilingual WebNLG Challenge}, Prague, Czech
  Republic.

\bibitem[{Papineni et~al.(2002)Papineni, Roukos, Ward, and
  Zhu}]{papineni2002bleu}
Kishore Papineni, Salim Roukos, Todd Ward, and Wei-Jing Zhu. 2002.
\newblock Bleu: a method for automatic evaluation of machine translation.
\newblock In \emph{Proceedings of the 40th annual meeting of the Association
  for Computational Linguistics}, pages 311--318.

\bibitem[{Popovi{\'c}(2017)}]{popovic-2017-chrf}
Maja Popovi{\'c}. 2017.
\newblock \href {https://doi.org/10.18653/v1/W17-4770} {chr{F}++: words helping
  character n-grams}.
\newblock In \emph{Proceedings of the Second Conference on Machine
  Translation}, pages 612--618, Copenhagen, Denmark. Association for
  Computational Linguistics.

\bibitem[{Snover et~al.(2006)Snover, Dorr, Schwartz, Micciulla, and
  Makhoul}]{snover-etal-2006-study}
Matthew Snover, Bonnie Dorr, Rich Schwartz, Linnea Micciulla, and John Makhoul.
  2006.
\newblock \href {https://aclanthology.org/2006.amta-papers.25} {A study of
  translation edit rate with targeted human annotation}.
\newblock In \emph{Proceedings of the 7th Conference of the Association for
  Machine Translation in the Americas: Technical Papers}, pages 223--231,
  Cambridge, Massachusetts, USA. Association for Machine Translation in the
  Americas.

\bibitem[{Wei et~al.(2022)Wei, Wang, Schuurmans, Bosma, Chi, Le, and
  Zhou}]{wei2022chain}
Jason Wei, Xuezhi Wang, Dale Schuurmans, Maarten Bosma, Ed~Chi, Quoc Le, and
  Denny Zhou. 2022.
\newblock Chain of thought prompting elicits reasoning in large language
  models.
\newblock \emph{arXiv preprint arXiv:2201.11903}.

\end{thebibliography}
\bibliographystyle{acl_natbib}

\appendix

\section{Prompts}
\label{sec:appendix_prompts}

We provide all the prompts used in our experiments, specifying the template and one example for each prompt. In Table \ref{tab:zero-shot} we provide the zero-shot minimal instruction prompt, while in Table \ref{tab:few-shot} we provide the few-shot in-context prompt with the same set of examples we used in our experiments.

\begin{table*}[h!]
\centering\small
\renewcommand{\arraystretch}{1.15}
\begin{tabular}{|>{\raggedright\arraybackslash}p{0.15\textwidth}|>{\raggedright\arraybackslash}p{0.75\textwidth}|}
\hline
\multicolumn{2}{|c|}{\textbf{Zero-Shot minimal instruction}} \\
\hline
\textbf{Template:} & Write the following triples as fluent {English | Irish | Welsh | Maltese | Breton} text.\bigbreak Triples: """\par \{set of triples in the format \textit{subject predicate object} and each triple in a new line\}\par """\bigbreak Text: [MODEL] \\
\hline
\textbf{Example Prompt:} & Write the following triples as fluent English text.\bigbreak Triples: """\par AC\_Hotel\_Bella\_Sky\_Copenhagen owner Bella\_Center \par AC\_Hotel\_Bella\_Sky\_Copenhagen tenant Marriott\_International \par AC\_Hotel\_Bella\_Sky\_Copenhagen architect 3XN \par AC\_Hotel\_Bella\_Sky\_Copenhagen floorCount 23\par """\bigbreak Text: \\
\textbf{Model output:} & \textit{The AC Hotel Bella Sky Copenhagen is owned by Bella Center, and is rented by Marriott International. It was designed by 3XN and has 23 floors.} \\
\hline
\end{tabular}
\caption{\label{tab:zero-shot}
Zero-Shot minimal instruction prompt. Template of the prompt and complete example in English.}
\end{table*}

\begin{table*}[h!]
\centering\small
\renewcommand{\arraystretch}{1.15}
\begin{tabular}{|>{\raggedright\arraybackslash}p{0.15\textwidth}|>{\raggedright\arraybackslash}p{0.75\textwidth}|}
\hline
\multicolumn{2}{|c|}{\textbf{Few-Shot in-context}} \\
\hline
\textbf{Template:} & Write the following triples as fluent {English | Irish | Welsh | Maltese | Breton} text.\bigbreak Triple 1: """\par \{set of triples in the format \textit{subject predicate object} and each triple in a new line\}\par """\par Text 1: \{verbalisation of Triple 1\}\par \#\#\par Triple 2: """\par \{set of triples in the format \textit{subject predicate object} and each triple in a new line\}\par """\par Text 2: \{verbalisation of Triple 2\}\par \#\#\par Triple 3: """\par \{set of triples in the format \textit{subject predicate object} and each triple in a new line\}\par """\par Text 3: [MODEL] \\
\hline
\textbf{Example Prompt:} & Write the following triples as fluent English text.\bigbreak Triple 1: """\par Adolfo\_Suárez\_Madrid–Barajas\_Airport runwayName "14R/32L"\par """\par Text 1: 14R/32L is the runway name of Adolfo Suárez Madrid-Barajas Airport.\par \#\#\par Triple 2: """\par American\_Journal\_of\_Mathematics abbreviation "Am. J. Math." \par American\_Journal\_of\_Mathematics firstPublicationYear 1878 \par American\_Journal\_of\_Mathematics issnNumber "1080-6377"\par """\par Text 2: The American Journal of Mathematics was first published in 1878 and is also known by the abbreviated title of Am. J. Math. It has an ISSN number of 1080-6377.\par \#\#\par Triple 3: """\par AC\_Hotel\_Bella\_Sky\_Copenhagen owner Bella\_Center \par AC\_Hotel\_Bella\_Sky\_Copenhagen tenant Marriott\_International \par AC\_Hotel\_Bella\_Sky\_Copenhagen architect 3XN \par AC\_Hotel\_Bella\_Sky\_Copenhagen floorCount 23 \par """\par Text 3: \\
\textbf{Model output:} & \textit{The AC Hotel Bella Sky Copenhagen is owned by Bella Center and is tenanted by Marriott International. It was designed by 3XN and has 23 floors.} \\
\hline \hline
\textbf{Irish texts:} & Text 1: 14R/32L is ainm do rúidbhealach Aerfort Adolfo Suárez Madrid-Barajas \par Text 2: Foilsíodh an American Journal of Mathematics don chéad uair in 1878 agus aithnítear leis an ainm giorraithe Am. J. Math. chomh maith é. Tá an uimhir ISSN 1080-6377 aige. \\
\hline
\textbf{Breton texts:} & Text 1: Anv leurenn bradañ aerborzh Adolfo Suárez Madrid-Barajas zo 14L/32R. \par Text 2: Finland zo bro ar Finniz hag hini ar skorndorrer Aleksey Chirikov bet savet e chanter-bigi Arctech en Helsinki. \\
\hline \hline  
\textbf{Maltese and Welsh Triples:} & Triple 1: Albennie\_Jones birthPlace Errata,\_Mississippi \bigbreak Triple 2: GMA\_New\_Media industry Entertainment\par GMA\_New\_Media type Media\_company\par GMA\_New\_Media product World\_Wide\_Web \\
\hline
\textbf{Maltese texts:} & Text 1: Albennie Jones twieldet f'Errata Mississippi. \par Text 2: GMA New Media hija kumpanija tal-midja tal-industrija tad-divertiment li toffri servizzi li jikkonċernaw il-World Wide Web. \\
\hline
\textbf{Welsh texts:} & Text 1: Ganed Albennie Jones yn Errata, Mississippi. \par Text 2: Mae GMA New Media yn gwmni cyfryngau yn y diwydiant adloniant sy'n cynnig gwasanaethau sy'n ymwneud â'r We Fyd Eang. \\
\hline
\end{tabular}
\caption{\label{tab:few-shot}
Few-Shot In-context prompt. \textbf{Top} Template and complete example in English. \textbf{Center} Examples' texts in Irish and Breton. \textbf{Bottom} Examples' triple set and texts in Maltese and Welsh.}
\end{table*}

\end{document}